# Few-shot Learning for Named Entity Recognition in Medical Text

**Maximilian Hofer**
University of Oxford
`maximilian.hofer@cs.ox.ac.uk`

**Andrey Kormilitzin**
University of Oxford
`andrey.kormilitzin@psych.ox.ac.uk`

**Paul Goldberg**
University of Oxford
`paul.goldberg@cs.ox.ac.uk`

**Alejo Nevado-Holgado**
University of Oxford
`alejo.nevado-holgado@psych.ox.ac.uk`

## Abstract

Deep neural network models have recently achieved state-of-the-art performance gains in a variety of natural language processing (NLP) tasks (Young, Hazarika, Poria, & Cambria, 2017). However, these gains rely on the availability of large amounts of annotated examples, without which state-of-the-art performance is rarely achievable. This is especially inconvenient for the many NLP fields where annotated examples are scarce, such as medical text. To improve NLP models in this situation, we evaluate five improvements on named entity recognition (NER) tasks when only ten annotated examples are available: (1) layer-wise initialization with pre-trained weights, (2) hyperparameter tuning, (3) combining pre-training data, (4) custom word embeddings, and (5) optimizing out-of-vocabulary (OOV) words. Experimental results show that the F1 score of 69.3% achievable by state-of-the-art models can be improved to 78.87%.

## 1  Introduction

Electronic health records (EHRs) are the databases used by general practitioners (GPs) and hospitals to build and store the medical history of patients (O. A. Johnson, Fraser, Wyatt, & Walley, 2014). These records include information such as the reason for administering drugs, previous disorders of the patient or the outcome of past treatments, and they are the largest source of empirical data in biomedical research, allowing for major scientific findings in highly relevant disorders such as cancer and Alzheimer's disease (Perera, Khondoker, Broadbent, Breen, & Stewart, 2014). However, most of the information held in these EHRs is in the form of natural language, making it largely inaccessible for statistical analysis (Murdoch & Detsky, 2013). Unlocking this information can bring a significant advancement to biomedical research.

Rule-based systems can extract medical information with good accuracy in simpler situations, such as when the to-be-extracted information follows regular speech patterns (e.g. mentions of regulated medical codes such as ICD9) (Karystianis et al., 2018). However, these systems don't scale well to complex patterns (e.g. descriptions of symptoms), variations of text patterns (e.g. American English against British English) or badly structured text (e.g. not standardized abbreviations are very common in EHRs), which is more akin the situation found in EHRs. Further, designing rule-based systems is very time-consuming and requires expert field knowledge. In these more complex situations, machine learning (ML) -based methods outperform rule-based ones by tuning general algorithms with existing data. Most recently, neural networks are being especially successful in complex NLP tasks (Young et al., 2017), where more traditional rule-based and other ML-based methods fail (Cambria & White, 2014). For instance, a combination of the long short-term memory (LSTM) type of recurrent neural networks (RNNs) and a convolutional neural network (CNN) has been successfully applied to set new state-of-the-art performance for NER tasks based on CoNLL-2003 and OntoNotes 5.0 data (J. Chiu & Nichols, 2016). The main remaining limitation of neural networks lies on the need for large amounts of annotated text, which is especially troublesome for their application to EHRs. Namely, only a few publicly available datasets exist for medical NLP, and even fewer exist with annotations for NLP tasks (e.g. document classification; or slot filling). The best known of these datasets are MIMIC-III (A. E. W. Johnson et al., 2016) and i2b2 (i2b2, 2018, p. 2). Therefore, improving the performance of neural networks when very few annotated examples are available remains a high priority in biomedical research.

| Dataset | NER category | Label count |
|---|---|---|
| i2b2 2009 [restricted] | Medication names | 236 |
| | Dosages | 134 |
| | Modes | 104 |
| | Frequencies | 116 |
| | Durations | 12 |
| | Reasons | 44 |
| i2b2 2010 | Tests | 2,513 |
| | Problems | 4,197 |
| | Treatments | 2,771 |
| i2b2 2012 | Problems | 2,989 |
| | Tests | 1,604 |
| | Treatments | 2,269 |
| | Clinical dep | 583 |
| | Evidentials | 447 |
| | Occurrences | 2,043 |
| CoNLL-2003 | Organizations | 6,321 |
| | Persons | 6,600 |
| | Locations | 7,140 |
| | Miscellaneous | 3,438 |

*Table 1. Data. Count of samples per named entity in i2b2 datasets used for supervised training. I2b2 2009 still contains further samples from 10 more records, which were only used for testing.*

In this paper, we demonstrate the effect of five sequential improvements on the learning capabilities of a neural network when having very few annotated examples. We start by setting the objective of optimizing performance on the NER task of i2b2 2009 while using only 10 randomly selected annotated discharge summaries. With a state-of-the-art NER architecture as the baseline (J. Chiu & Nichols, 2016), we sequentially design and apply a number of improvements that substantially improve performance on this objective. This improves performance from 69.3% at baseline to 78.87%.

## 2 Methods

### 2.1. Data

Six medical NER datasets were used: one defining the target task and therefore used for supervised training and testing (i2b2 2009), two for supervised pre-training of weights (i2b2 2010 and i2b2 2012), and three for unsupervised training of custom word embeddings (BioNLP-2016, MIMIC-III and UK CRIS). In addition, a non-medical dataset (CoNLL-2003) was also used for supervised pre-training of weights. Each of the datasets used in a supervised fashion (i.e. i2b2 sets and CoNLL-2003) provided a number of target NER categories that were applied as labels (see table 1), while in the datasets used in an unsupervised fashion original annotations were ignored (see table 2). We have restricted the i2b2b 2009 data to only 10 random samples drawn from the full training dataset.

The Informatics for Integrating Biology & the Bedside (i2b2) non-profit foundation has run a series of yearly NLP challenges, publishing digital copies of annotated, hand-written and fully deidentified clinical notes. The i2b2 challenge 2009 focused on extracting medication information from de-identified discharge summaries from Partners Healthcare (O. Uzuner, Solti, & Cadag, 2010). The 2010 i2b2/VA Relations Challenge used discharge summaries and progress reports from Partners Healthcare, Beth Israel Deaconess Medical Center, and the University of Pittsburgh Medical Center (Ö. Uzuner, South, Shen, & DuVall, 2011). The 2012 i2b2 challenge included a task on extracting problems, tests, treatments, clinical departments, evidentials (i.e. events indicating the source of information) and occurrences (e.g. admission, transfer) in discharge summaries provided by Partners Healthcare and the Beth Israel Deaconess Medical Center (Sun, Rumshisky, & Uzuner, 2013).

The Conference on Natural Language Learning (CoNLL) is a yearly conference organised by SIGNLL (ACL's Special Interest Group on Natural Language Learning). CoNLL-2003 includes English language newswire articles, annotated with persons, locations, organizations and names of miscellaneous entities that do not belong to any of the previous three groups (Sang & De Meulder, 2003).

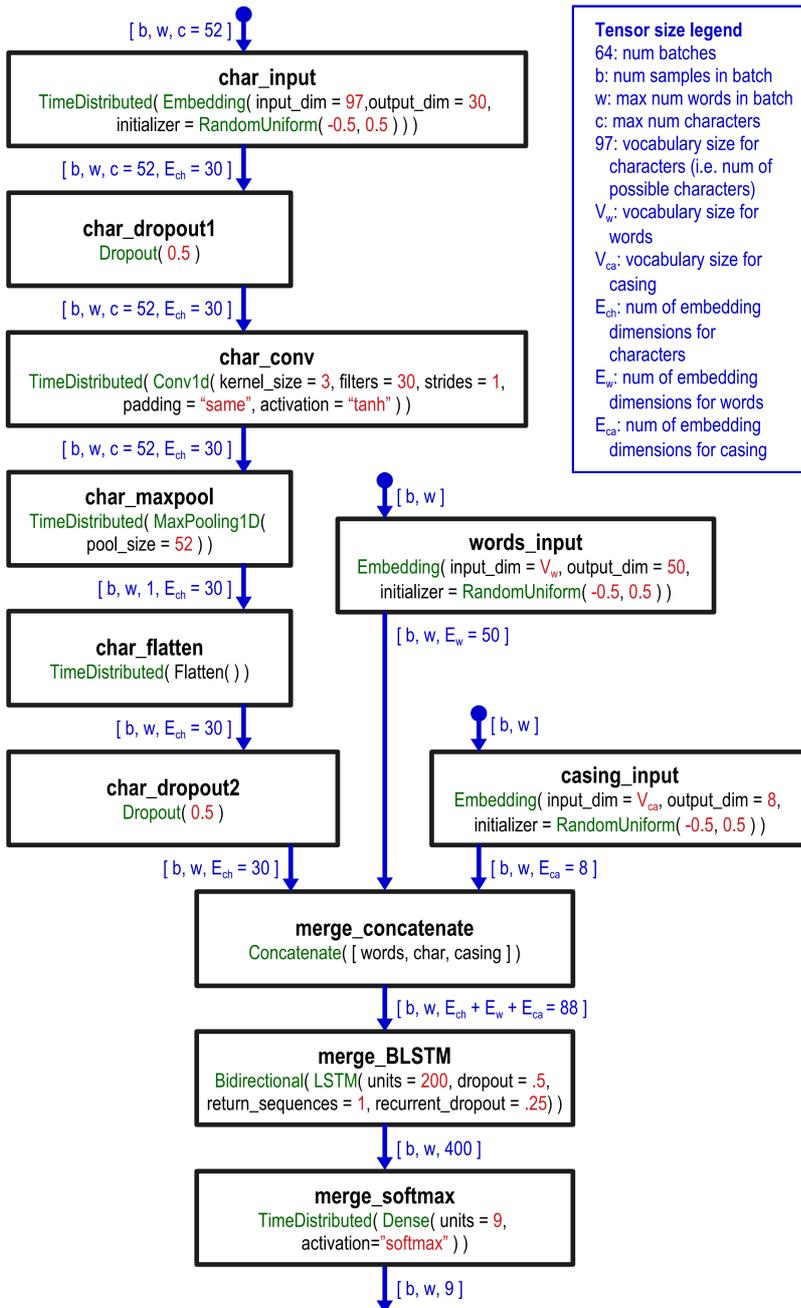

*Figure 1. Model architecture. Operations are represented by rectangles, with operation name in black bold letters, its corresponding Keras code below, function names in green and non-default parameters in red. Tensors exchanged by operations are represented by blue arrows, with the size of each tensor in brackets. A legend in the upper right corner lists the meaning of some numbers and variables used in the diagram*

Deaconess Medical Center in Boston, Massachusetts (A. E. W. Johnson et al., 2016).

UK CRIS allows controlled access to EHRs of mental health hospitals across the UK, among which we accessed the Oxford Mental Health Clinic at Warneford Hospital, containing approximately 500,000 patient records (Callard et al., 2014; Stewart et al., 2009). This dataset is maintained by the UK National Health Service (NHS), and is only accessible through controlled ethical approval and audit processes (further information at https://crisnetwork.co/).

### 2.2. Baseline model

The baseline model is based on the state-of-the-art NER architecture for CoNLL-2003 and OntoNotes 5.0 text data (J. Chiu & Nichols, 2016). The model has three inputs, namely a character-level, word-level and casing input, each of which encodes a different aspect of the text (see Figure 1). The architecture starts processing these three inputs independently, but then merges them to process further. This architecture does so through a number of atomic operations or layers, which we describe as follows:

- **Character embedding** layer (char_input in Figure 1) maps a vocabulary of 97 possible characters to 30-dimensional embedding, initialized randomly from $U(-0.5, 0.5)$. The number of input samples per batch ('b' in Figure 1) and the number of words per sample ('w') vary from batch to batch. The maximum number of characters per word ('c') was 52.
- **Dropout** layers (char_dropout1 and char_dropout2 in Figure 1) with drop rate 0.5 are applied to the character-level input to mitigate the risk of overfitting.

BioNLP-2016 provides word vectors based on text data from a PubMed Central Open Access subset (PMC) and PubMed (B. Chiu, Crichton, Korhonen, & Pyysalo, 2016).

The MIMIC III v1.4 database consists of 58,000 critical care hospital notes from Beth Israel

- **1D convolutional** layer (char_conv) processes the 1-dimensional character input with 30 kernels of width 3. This layer is followed by a 1d maxpool operation (char_maxpool) of window size 52 and stride of 52, which effectively compiles the character dimension into size 1. The kernel is initialized by drawing from a Glorot uniform distribution (Glorot & Bengio, 2010). Bias terms are initialized to zero.
- **Word embedding** layer (words_input) maps a vocabulary of '$V_w$' words into 50-dimensional embeddings. Unless stated otherwise, we use the GloVE Wikipedia 2014 and Gigaword 5 embeddings with 6B tokens (Pennington, Socher, & Manning, 2014).
- **Casing embedding** layer (casing_input) maps a vocabulary of '$V_{ca}$' casing types into $V_{ca}$-dimensional embeddings. By default, 8 casing types are considered: *numeric*, *allLower*, *allUpper*, *mainly_numeric* (more than 50% of characters of a word are numeric), *initialUpper*, *contains_digit*, *padding* and *other* (if no category was applicable).
- **Concatenation** layer (merge_concatenate) combines processed character-level (a vector of 30 dimensions per sample input), word-level (50 dimensions) and casing ($V_{ca}$ dimensions) data into a vector of 80 + $V_{ca}$ dimensions.
- **Bidirectional LSTM (BLSTM)** (Schuster & Paliwal, 1997) layer (merge_BLSTM) transforms the previously concatenated data into two vectors of 200 units, one applying forwards and another backwards recursion on the input The kernels are initialized by drawing from a Glorot uniform distribution (Glorot & Bengio, 2010). Bias terms are initialized to zero.
- **Dense output layer** (merge_softmax) applies a layer-wise softmax function to output a prediction for locating and classifying sequences of words in the input text. The number of units depends on the specific objective task. The kernel is initialized by drawing from a Glorot uniform distribution (Glorot & Bengio, 2010). Bias terms are initialized to zero.

In this baseline model, all parameters are trained with Nadam optimizer with default parameters (as defined in Keras version 2.2.0), dividing data into 64 batches. The baseline model is implemented in Python using Keras and TensorFlow libraries and available from https://github.com/mxhofer/Named-Entity-Recognition-BidirectionalLSTM-CNN-CoNLL.

### 2.3. Single pre-training

An approach that is generally successful in computer vision tasks where few annotated examples are available consists of pre-training on another related task where substantially more samples are available. For example, one study pre-trained on a large labeled training corpus of still images and successfully transferred learning to a more sparsely labelled image corpus for video recognition (Su, Chiu, Yeh, Huang, & Hsu, 2014). This pre-training approach is the first method we applied to improve the baseline architecture performance. Network parameters were pre-trained separately on each of three distinct NER tasks, two of them belonging to the same domain as the target task (i2b2 2010 and i2b2 2012, medical text), and one belonging to a different domain (CoNLL-2003, non-medical text). Three different initialization strategies were compared: initializing all layers with pre-trained weights; initializing only layer merge_BLSTM (other layers are initialized randomly); and initializing all but the merge_BLSTM (the merge_BLSTM is initialized randomly). In all cases, the embeddings of words_input are not trained but rather frozen to the values of GloVE.

### 2.4. Hyperparameter tuning

Our second improvement, incorporated in addition to this one described in section 3.3, consisted of tuning hyperparameter values via grid search. This is a common approach that has been successfully applied across multiple NLP tasks, such as WikiQA and SemEval-2016 (Min, Seo, & Hajishirzi, 2017). The hyperparameters that we fine-tuned were: optimizers (selecting either stochastic gradient descent (SGD) or Nadam); pre-training dataset (either i2b2 2010 or i2b2 2012); SGD learning rates (0.04 or 0.08); batch normalization (with or without); trainable word embeddings (weights of layer 'words_input' trained on the objective task or frozen to GloVE values) and learning rate decay (constant or time scheduled).

| Initialization | Layers | F1 score |
|---|---|---|
| Random | All | 69.30 |
| CoNLL-2003 | All | 68.47 |
| | BLSTM only | 71.19 |
| | All but BLSTM | 69.80 |
| i2b2 2010 | All | **73.82** |
| | BLSTM only | 72.02 |
| | All but BLSTM | 71.23 |
| i2b2 2012 | All | 71.32 |
| | BLSTM only | 70.42 |
| | All but BLSTM | 70.91 |

*Table 2. Single pre-training.* F1 scores per type of initialization (random, or pre-trained in CoNLL-2003, i2b2 2010 or i2b2 2012) and layer (all, BLSTM only or all but BLSTM). The best-performing combination is highlighted in bold.

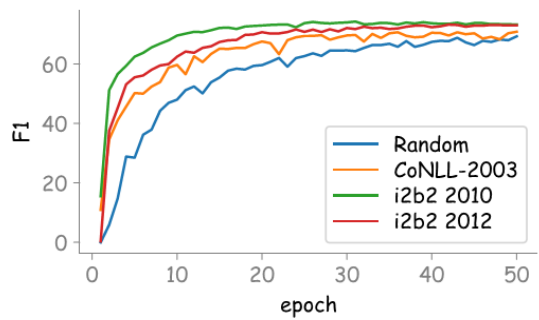

*Figure 2. Learning speed during single pre-training.* Aggregate F1 scores for different initializations (random, CoNLL-2003, i2b2 2010 and i2b2 2012) applied to all layers of the target domain, plotted over the number of training epochs.

### 2.5. Combined pre-training

Besides pre-training in individual datasets (as done in section 2.3), pre-training in combined datasets can further improve performance on the target task. For example, a health informatics study has used six publicly available datasets for lung CT scans to identify lung patterns using convolutional neural networks, which resulted in a 2% performance increase in the target domain (Christodoulidis, Anthimopoulos, Ebner, Christe, & Mougiakakou, 2017). To test whether our target task benefited from combined pre-training, now we combined learning from both i2b2 2010 and 2012. This is achieved by sequentially learning from each dataset in either of two possible directions: first training a randomly initialized model on i2b2 2010, and then continuing training on i2b2 2012; or first training on 2010's and then on 2012's. In either case, the final weights obtained from the second training round were then used as initial values when training in the objective task.

### 2.6. Customized word embeddings

While previously (sections 2.2, 2.3, 2.4 and 2.5) we used GloVE embeddings for the words input (word_input in Figure 1) these vector representations are expected to be inaccurate for medical terms, as they rarely appear in the general domain corpora used to create GloVE. To mend this problem, our fourth improvement consisted of developing our own word embeddings trained on either CRIS, MIMIC III or BioNLP-2016. CRIS and MIMIC III embeddings were trained with Facebook's FastText (Bojanowski, Grave, Joulin, & Mikolov, 2016; Joulin, Grave, Bojanowski, & Mikolov, 2016) algorithm, a minimum word count of 5, an initial learning rate of 0.05 and context window size of 5. BioNLP-2016 were downloaded from the official repository (Cambridge Language Technology Lab, 2018) and used without further preprocessing.

### 2.7. Optimizing OOV words

Data inspection revealed that the target dataset (i2b2 2009) contained a high proportion of out-of-vocabulary (OOV) words. These are words not included in the vocabulary of the embeddings that can have a highly detrimental impact on performance. In our case, a large proportion of OOV terms included trailing and leading characters, such as "increase dosage: +20 mgs week". To reduce OOV, our last improvement added the following two steps to text preprocessing:

- Remove trailing ":", ";", "." and "-".
- Remove quotations
- Remove leading "+"

## 3 Results

As described in the introduction, first we defined a few shots learning objective task, which consisted on the official objective of the i2b2 challenge of 2009 but being allowed to train only on 10 annotated, randomly sampled discharge summaries. Subsequently, we implemented one of the state of the art NER architectures and evaluated the effect of five sequential improvements over this baseline model.

### 3.1. Baseline model

The baseline model is based on a state-of-the-art NER architecture initially proposed for CoNLL-2003 and OntoNotes 5.0 corpora (J. Chiu & Nichols, 2016). As described in methods (see Figure 1), it consists of a BLSTM with casing, word and character level inputs, with the latter one also undergoing one convolution and dropout. Layer 'words_input' is initialized with GloVE embeddings trained on Wikipedia 2014 and Gigaword (6B tokens). These embeddings are frozen after initialization and not modified further during backpropagation. Embeddings of 'char_input' and 'casing_input' are randomly initialized with the uniform distribution $U(-0.5, 0.5)$. All other parameters are randomly initialized following Keras (version 2.2.0) defaults. Data is then divided into 64 batches and all not frozen embeddings, weights and biases trained by Nadam with default Keras parameters. This baseline architecture with this initialization and training method obtained an F1 score of 69.30% on our objective task (see Figure 2).

### 3.2. Single pre-training

Single pre-training is the first improvement implemented over the baseline model. While the baseline model initialized all parameters randomly (except the embeddings of layer 'words_input'), single pre-training initialized either parts or the entire neural network by pre-training in other datasets. Initializing layers by pre-trained on i2b2 2010 (average F1 increase of +3.06% over baseline) or i2b2 2012 (+1.58%) performed better than weights pre-trained on CoNLL-2003 (+0.52%) and better than randomly initialized weights (69.3%). Furthermore, initializing all layers performs both better than initializing only the BLSTM (71.21%) or all but the BLSTM layer (70.65%). Results of all combinations are shown in Table 2. Pre-training all layers and on i2b2 2010 was the combination achieving the best performance with an F1 score of +4.52% with respect to the baseline. We, therefore, incorporated this improvement into the model for subsequent experiments.

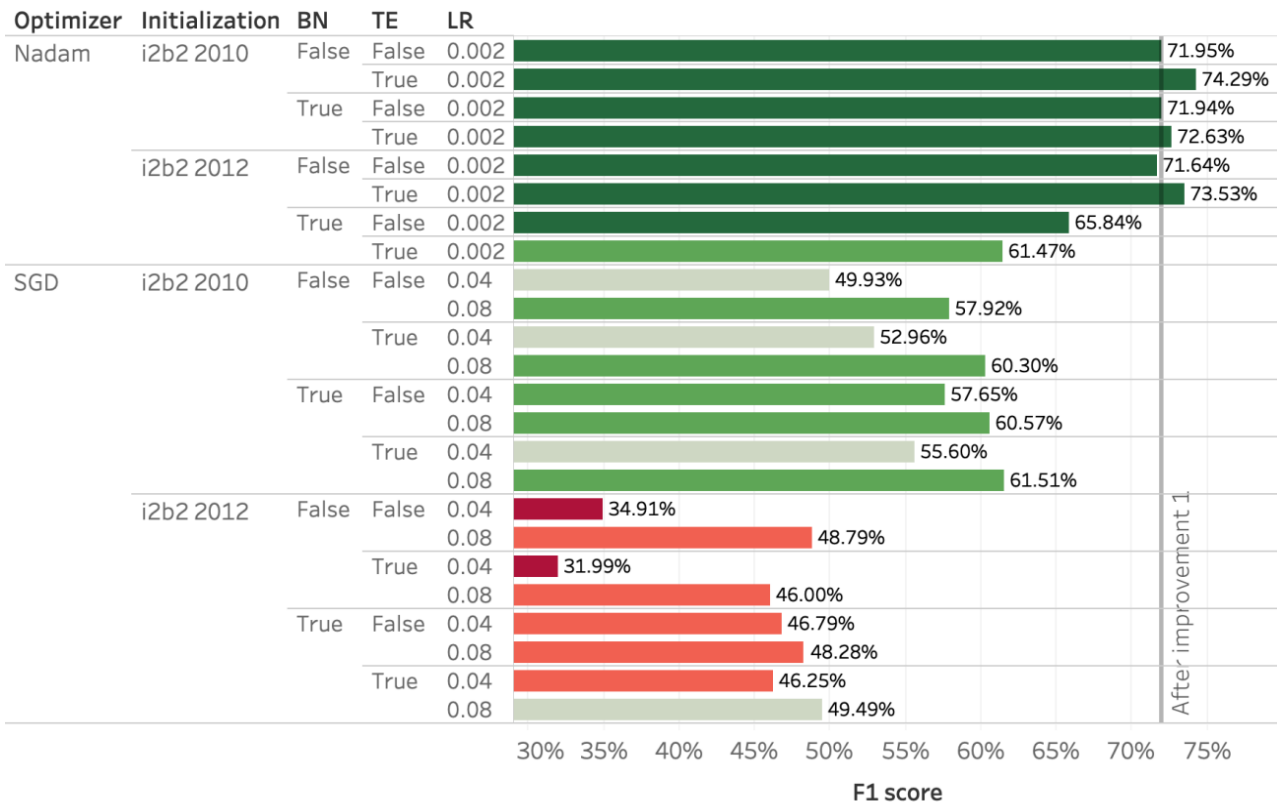

*Figure 3. Tuning hyperparameters.* F1 scores split by optimizers, initializations, batch normalization (BN), trainable embeddings (TE) and learning rates (LR). All SGD learning rates are constant.

Interestingly, pre-training on datasets of the medical domain enabled faster learning than otherwise. By epoch 5, the average F1 score of i2b2 2010 and 2012 was 58.98%, but 50.2% for CoNLL-2003 and 28.47% for random initializations (see Figure 2).

### 3.3. Hyperparameters

Tuning hyperparameters is the second improvement tested, and it was implemented over the best performing model with single pre-training. Of all the evaluated hyperparameters, the one with the largest impact was the optimizer, with Nadam achieving 70.41% on average and SGD 50.56%. The second most important hyperparameter was the data used for pre-training the neural network, with i2b2 2010 achieving up to +2.34% and i2b2 2012 up to +1.58%. The effects of other hyperparameters (batch normalization, trainable embeddings, learning rates and learning rate decay) were inconclusive, as visualized in Figure 3. We, therefore, fixed the optimizer to Nadam and kept using i2b2 2010 for pre-training, while rejecting other hyperparameter changes.

### 3.4. Combined pre-training

Pre-training simultaneously with several datasets is the third improvement tested, and it was implemented over the best model of fine-tuned hyperparameters. The results show that separate pre-training with i2b2 2010 outperforms combined pre-training of either direction. Pre-training first on 2010's and then on 2012's obtained -1.85% F1 while the opposite direction resulted on -1.66%. We, therefore, rejected combined pre-training and continued using i2b2 2010 alone for subsequent experiments.

### 3.5. Customizing word embeddings

Customizing the embeddings of layer 'words_input' by pre-training them is the fourth improvement tested, and it was implemented over the best model of fine-tuned parameters (combined pre-training was rejected). All previous models were using GloVE word embeddings with 50 dimensions, with values fixed to those publicly available

| Embedding | Dim. | Initializat. | F1 |
|---|---|---|---|
| GloVE | 50 | i2b2 2010 | 72.29 |
| GloVE | 50 | i2b2 2012 | 72.13 |
| MIMIC III | 50 | i2b2 2010 | 73.34 |
| MIMIC III | 50 | i2b2 2012 | 72.21 |
| MIMIC III | 200 | i2b2 2010 | **78.07** |
| MIMIC III | 200 | i2b2 2012 | 76.91 |
| BioNLP-2016 2w | 200 | i2b2 2010 | 66.89 |
| BioNLP-2016 2w | 200 | i2b2 2012 | 66.76 |
| BioNLP-2016 30w | 200 | i2b2 2010 | 66.25 |
| BioNLP-2016 30w | 200 | i2b2 2012 | 64.64 |
| CRIS | 50 | i2b2 2010 | 68.24 |
| CRIS | 50 | i2b2 2012 | 68.75 |
| CRIS | 200 | i2b2 2010 | 72.13 |
| CRIS | 200 | i2b2 2012 | 71.02 |

*Table 3. Customizing word embeddings.* F1 scores for each word embedding, word vector dimension and i2b2 initialization. The best-performing combination is highlighted in bold. 'Dim' stands for 'dimentions', 'Initializat' for 'initialization'.

(https://nlp.stanford.edu/projects/glove/). Replacing these with embeddings trained on MIMIC III (word vectors of 50 or 200 dimensions) achieved +3.78% F1 score (200 dimensions) with respect to the best result of section 3.3; BioNLP-2016 (200 dimensions, window size, *w*, of 2 or 30) -7.4%; and UK CRIS (50 or 200 dimensions) -2.16%. For completion, all three possibilities were evaluated while pre-training the rest of the network in either i2b2 2010 or i2b2 2012 (see Table 3 for the result of each combination). The best combination was MIMIC III with i2b2 2010 with an F1 score of 78.07% (+3.78%).

Interestingly, MIMIC III was also the dataset with fewest OOV terms with respect to the objective task (see Table 4).

### 3.6. Optimizing OOV words

Reducing the number of OOV words in the embeddings of layer 'words_input' was the fifth and final improvement tested, and it was implemented over the model with best-customized embeddings. The number of OOV words depended on the dataset used to generated the word embeddings (see Table 4). The additional preprocessing steps described in section 3.7 reduced OOV words by 3% for MIMIC III, 7% for CRIS, 11% for GloVE and 11% for BioNLP-2016. The results show that this step can improve the F1 score marginally in all cases, but never decreases performance. Using i2b2 2010 for pre-training weights, MIMIC III embeddings and the additional pre-processing steps increase the F1 score from 78.07% to 78.87%.

| Embedding | OOV terms | Decrease after optimization |
|---|---|---|
| GloVE | 3,583 | 11% |
| BioNLP-2016 | 3,537 | 11% |
| CRIS-2003 | 5,838 | 7% |
| MIMIC III | 3,090 | 3% |

*Table 4. OOV terms.* Number of OOV terms or each word embedding.

## 4 Discussion

The problem of having limited annotated samples has been addressed in several domains with one-shot (Fei-Fei, Fergus, & Perona, 2006; Vinyals, Blundell, Lillicrap, Kavukcuoglu, & Wierstra, 2016) and even zero-shot learning (Xian, Lampert, Schiele, & Akata, 2017). However, to our knowledge, this challenge has not been addressed in medical text, where the problem is especially taxing. Namely, medical text from EHRs is different from other corpora, to the point that non-clinically trained readers can hardly understand EHR text due to the extensive use of technical terms, non-standard acronyms and unofficial shorthand. For this reason, standard transfer learning methods from other better annotated but non-medical corpora (e.g. using standard Glove embeddings or language models trained in Wikipedia) are expected to perform badly. Hence, we believe that contributions such as this paper are especially important.

Besides the final model, our experiments make a number of observations applicable to this challenge. The results derived from our first improvement (section 3.2) over the baseline model suggest that pre-trained weights based on text similar to the target domain substantially improve performance (see Table 2 and Figure 2). Better transfer learning within domain than across domain has also been reported elsewhere, e.g. (Tan, Zhang, Pan, & Yang, 2017). The improvement we observe is especially drastic for the first stages of learning and tends to diminish towards an asymptotic positive constant after long learning (see Figure 2).

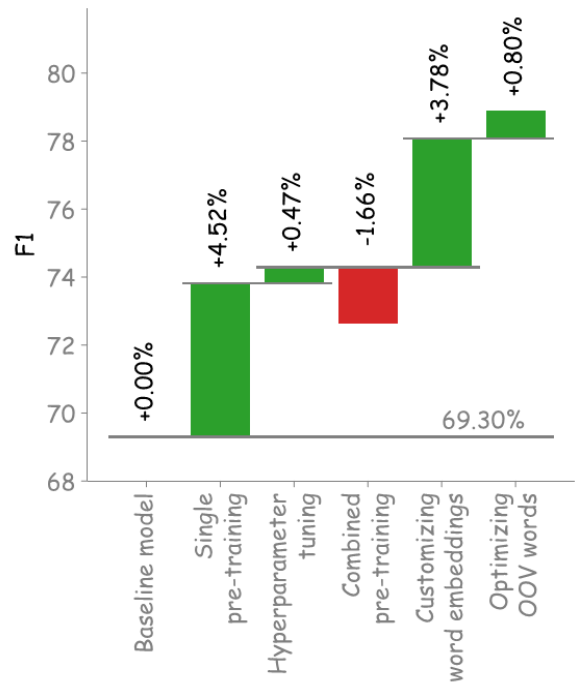

*Figure 4. Summary.* The figure shows the increase (green) or decrease (red) in performance (y-axis) produced after each improvement (x-axis). Baseline F1 score is 69.3% and the final F1 score is 78.87%.

Our second model improvement (section 3.3) indicates that the choice of optimizer has a larger effect on the F1 score than using batch normalization (BN) and trainable word embeddings (TE). BN and TE show inconclusive results, while Nadam optimizer outperforms SGD in all the situations tested (see Figure 3). This is in line with a recent study that has evaluated the impact of different hyperparameters for NER tasks. They have found that the Nadam optimizer performs best while also converging the fastest (Reimers & Gurevych, 2017).

Combining pre-training data, our third improvement, decreased performance (section 3.4). There are two possible explanations for this. Firstly, we have optimized hyperparameters on separate datasets in section 3.3. Secondly, when the model has trained on the first dataset and then refines the weights on the second dataset, it might find the initial values unsuitable and subsequently has to adjust them by a lot more than if the weights were initialized randomly, which is the case for generating separate pre-trained weights.

Finally, results from the fourth improvement (section 3.5) suggest that custom word embeddings trained on text from the same domain as the objective task specially improve performance (see Table 3). This may be due in part to the fewer OOV words resulting from embeddings training on domain-specific corpora (see Table 4).

As a final comment, our study still has several limitations, and future work should address these. Firstly, the improvement steps were applied sequentially, so a different ordering could further improve performance on the objective task. Given this search space limitation, the results found in this study can be considered conservative. Finally, an F1 score of 78.87%, is still fairly low compared to the performance achievable with large annotated corpora.

## Acknowledgements

UK CRIS was established and receives continuing funding from the National Institute for Health Research (NIHR) and the Medical Research Council (MRC), including through Dementias Platform UK, and from the NIHR Biomedical Research Centre at Oxford Health National Health Service (NHS) Foundation Trust and the University of Oxford.

## References

Bojanowski, P., Grave, E., Joulin, A., & Mikolov, T. (2016). Enriching Word Vectors with Subword Information. *ArXiv:1607.04606 [Cs]*. Retrieved from http://arxiv.org/abs/1607.04606

Callard, F., Broadbent, M., Denis, M., Hotopf, M., Soncul, M., Wykes, T., … Stewart, R. (2014). Developing a new model for patient recruitment in mental health services: a cohort study using Electronic Health Records. *BMJ Open*, *4*(12), e005654. https://doi.org/10.1136/bmjopen-2014-005654

Cambria, E., & White, B. (2014). Jumping NLP Curves: A Review of Natural Language Processing Research [Review Article]. *Comp. Intell. Mag.*, *9*(2), 48–57. https://doi.org/10.1109/MCI.2014.2307227

Cambridge Language Technology Lab. (2018). *GitHub repository cambridgeltl/BioNLP-2016*. Python, Cambridge Language Technology Lab. Retrieved from https://github.com/cambridgeltl/BioNLP-2016

Chiu, B., Crichton, G., Korhonen, A., & Pyysalo, S. (2016). How to Train good Word Embeddings for Biomedical NLP - Dimensions. In *Proceedings of the 15th Workshop on Biomedical Natural Language Processing*. Retrieved from https://app.dimensions.ai/details/publication/pub.1098653399

Chiu, J., & Nichols, E. (2016). Named Entity Recognition with Bidirectional LSTM-CNNs. *ArXiv:1511.08308 [Cs]*. Retrieved from http://arxiv.org/abs/1511.08308

Christodoulidis, S., Anthimopoulos, M., Ebner, L., Christe, A., & Mougiakakou, S. (2017). Multisource Transfer Learning With Convolutional Neural Networks for Lung Pattern Analysis. *IEEE Journal of Biomedical and Health Informatics*, *21*(1), 76–84. https://doi.org/10.1109/JBHI.2016.2636929

Fei-Fei, L., Fergus, R., & Perona, P. (2006). One-shot learning of object categories. *IEEE Transactions on Pattern Analysis and Machine Intelligence*, *28*(4), 594–611. https://doi.org/10.1109/TPAMI.2006.79

Glorot, X., & Bengio, Y. (2010). Understanding the difficulty of training deep feedforward neural networks. In *Proceedings of the Thirteenth International Conference on Artificial Intelligence and Statistics* (pp. 249–256). Retrieved from http://proceedings.mlr.press/v9/glorot10a.html

i2b2. (2018, August 8). i2b2: Informatics for Integrating Biology & the Bedside. Retrieved 15 July 2018, from https://www.i2b2.org/

Johnson, A. E. W., Pollard, T. J., Shen, L., Lehman, L. H., Feng, M., Ghassemi, M., … Mark, R. G. (2016). MIMIC-III, a freely accessible critical care database. *Scientific Data*, *3*, 160035. https://doi.org/10.1038/sdata.2016.35

Johnson, O. A., Fraser, H. S. F., Wyatt, J. C., & Walley, J. D. (2014). Electronic health records in the UK and USA. *The Lancet*, *384*(9947), 954. https://doi.org/10.1016/S0140-6736(14)61626-3

Joulin, A., Grave, E., Bojanowski, P., & Mikolov, T. (2016). Bag of Tricks for Efficient Text Classification. *ArXiv:1607.01759 [Cs]*. Retrieved from http://arxiv.org/abs/1607.01759

Karystianis, G., Nevado, A. J., Kim, C.-H., Dehghan, A., Keane, J. A., & Nenadic, G. (2018). Automatic mining of symptom severity from psychiatric evaluation notes. *International Journal of Methods in Psychiatric Research*, *27*(1), e1602. https://doi.org/10.1002/mpr.1602

Min, S., Seo, M., & Hajishirzi, H. (2017). Question Answering through Transfer Learning from Large Fine-grained Supervision Data. *ArXiv:1702.02171 [Cs]*. Retrieved from http://arxiv.org/abs/1702.02171

Murdoch, T. B., & Detsky, A. S. (2013). The inevitable application of big data to health care. *JAMA*, *309*(13), 1351–1352. https://doi.org/10.1001/jama.2013.393

Pennington, J., Socher, R., & Manning, C. (2014). Glove: Global Vectors for Word Representation. In *EMNLP* (Vol. 14, pp. 1532–1543). https://doi.org/10.3115/v1/D14-1162

Perera, G., Khondoker, M., Broadbent, M., Breen, G., & Stewart, R. (2014). Factors Associated with Response to Acetylcholinesterase Inhibition in Dementia: A Cohort Study from a Secondary Mental Health Care Case Register in London. *PLOS ONE*, *9*(11), e109484. https://doi.org/10.1371/journal.pone.0109484

Reimers, N., & Gurevych, I. (2017). Optimal Hyperparameters for Deep LSTM-Networks for Sequence Labeling Tasks. *ArXiv:1707.06799 [Cs]*. Retrieved from http://arxiv.org/abs/1707.06799

Sang, E. F. T. K., & De Meulder, F. (2003). Introduction to the CoNLL-2003 Shared Task: Language-Independent Named Entity Recognition. *ArXiv:Cs/0306050*. Retrieved from http://arxiv.org/abs/cs/0306050

Schuster, M., & Paliwal, K. K. (1997). Bidirectional recurrent neural networks. *IEEE Transactions on Signal Processing*, *45*(11), 2673–2681. https://doi.org/10.1109/78.650093

Stewart, R., Soremekun, M., Perera, G., Broadbent, M., Callard, F., Denis, M., … Lovestone, S. (2009). The South London and Maudsley NHS Foundation Trust Biomedical Research Centre (SLAM BRC) case register: development and descriptive data. *BMC Psychiatry*, *9*(1), 51. https://doi.org/10.1186/1471-244X-9-51

Su, Y.-C., Chiu, T.-H., Yeh, C.-Y., Huang, H.-F., & Hsu, W. H. (2014). Transfer Learning for Video Recognition with Scarce Training Data for Deep Convolutional Neural Network. *ArXiv:1409.4127 [Cs]*. Retrieved from http://arxiv.org/abs/1409.4127

Sun, W., Rumshisky, A., & Uzuner, O. (2013). Evaluating temporal relations in clinical text: 2012 i2b2 Challenge. *Journal of the American Medical Informatics Association: JAMIA*, *20*(5), 806–813. https://doi.org/10.1136/amiajnl-2013-001628

Tan, B., Zhang, Y., Pan, S. J., & Yang, Q. (2017). Distant Domain Transfer Learning. In *AAAI*.

Uzuner, O., Solti, I., & Cadag, E. (2010). Extracting medication information from clinical text. *Journal of the American Medical Informatics Association: JAMIA*, *17*(5), 514–518. https://doi.org/10.1136/jamia.2010.003947

Uzuner, Ö., South, B. R., Shen, S., & DuVall, S. L. (2011). 2010 i2b2/VA challenge on concepts, assertions, and relations in clinical text. *Journal of the American Medical Informatics Association: JAMIA*, 552–556. https://doi.org/10.1136/amiajnl-2011-000203

Vinyals, O., Blundell, C., Lillicrap, T., Kavukcuoglu, K., & Wierstra, D. (2016). Matching Networks for One Shot Learning. Retrieved from https://arxiv.org/abs/1606.04080

Xian, Y., Lampert, C. H., Schiele, B., & Akata, Z. (2017). Zero-Shot Learning - A Comprehensive Evaluation of the Good, the Bad and the Ugly. Retrieved from https://arxiv.org/abs/1707.00600

Young, T., Hazarika, D., Poria, S., & Cambria, E. (2017). Recent Trends in Deep Learning Based Natural Language Processing. Retrieved from https://arxiv.org/abs/1708.02709